\documentclass[10pt,twocolumn,letterpaper]{article}
\usepackage{cvpr}              % To produce the CAMERA-READY version
\usepackage{graphicx}
\usepackage{amsmath}
\usepackage{amssymb}
\usepackage{mathrsfs}
\usepackage{amsthm}

\usepackage{booktabs}
\usepackage{threeparttable}
\usepackage{multirow}
\usepackage{bbding}
\usepackage{color}
\usepackage[accsupp]{axessibility}

\makeatletter
\def\hlinewd#1{%
  \noalign{\ifnum0=`}\fi\hrule \@height #1 \futurelet
   \reserved@a\@xhline}
\makeatother

\usepackage[pagebackref,breaklinks,colorlinks]{hyperref}

\usepackage[capitalize]{cleveref}
\crefname{section}{Sec.}{Secs.}
\Crefname{section}{Section}{Sections}
\Crefname{table}{Table}{Tables}
\crefname{table}{Tab.}{Tabs.}

\def\tb{\textbf}

\def\confName{CVPR}
\def\confYear{2023}

\begin{document}

\title{Towards a Smaller Student: Capacity Dynamic  Distillation \\for Efficient Image Retrieval}

% \author{\IEEEauthorblockN{Yi Xie \textsuperscript{1} \quad 
% 		Huaidong Zhang \textsuperscript{1}\thanks{Corresponding author} 
%          \quad Xuemiao Xu \textsuperscript{1} 
%          \quad Jianqing Zhu  \textsuperscript{2} 
%          \quad Shengfeng He \textsuperscript{3}} \\
% 	\IEEEauthorblockA{
% 		\textsuperscript{1} South China University of Technology \quad
% 		\textsuperscript{2} Huaqiao University \quad
%   \textsuperscript{3} Singapore Management University}\\
% 	\IEEEauthorblockA{\tt\small ftyixie@mail.scut.edu.cn, \{huaidongz, xuemx\}@scut.edu.cn \\
% \tt\small jqzhu@hqu.edu.cn, shengfenghe@smu.edu.sg.}}

\author{%
Yi Xie$^{1}$ \quad Huaidong Zhang$^{1}$\thanks{Corresponding authors} \quad Xuemiao Xu$^{1,4,5,6*}$ \quad Jianqing Zhu$^{2}$  \quad Shengfeng He$^{3}$\\
$^{1}$South China University of Technology \; $^{2}$Huaqiao University \; $^{3}$Singapore Management University\\
$^{4}$State Key Laboratory of Subtropical Building Science \\ $^{5}$Ministry of Education Key Laboratory of Big Data and Intelligent Robot\\ $^{6}$Guangdong Provincial Key Lab of Computational Intelligence and Cyberspace Information\\
\tt\small ftyixie@mail.scut.edu.cn, \{huaidongz, xuemx\}@scut.edu.cn \\
 \tt\small jqzhu@hqu.edu.cn, shengfenghe@smu.edu.sg
}

\maketitle
%%%%%%%%% ABSTRACT
% \begin{abstract}
% Previous Knowledge Distillation based efficient image retrieval methods employs a lightweight network as the student model for fast inference. However, the lightweight student model lacks adequate representation capacity for effective knowledge imitation during the most critical early training period, causing final performance degeneration. To tackle this issue, we propose a Capacity Dynamic Distillation framework, which constructs a student model with editable representation capacity. Specifically, the employed student model is initially a heavy model to fruitfully learn distilled knowledge in the early training epochs, and the student model is gradually compressed during the training.
% To dynamically adjust the model capacity, our dynamic framework inserts a learnable convolutional layer within each residual block in the student model as the channel importance indicator. The indicator is optimized simultaneously by the image retrieval loss and the compression loss, and a retrieval-guided gradient resetting mechanism is proposed to release the gradient conflict. Extensive experiments show that our method has superior inference speed and accuracy, e.g., on the VeRi-776 dataset, given the ResNet101 as a teacher, our method saves 67.13\% model parameters and 65.67\% FLOPs (around 24.13\% and 21.94\% higher than state-of-the-arts) without sacrificing accuracy (around 2.11\% mAP higher than state-of-the-arts).
% \end{abstract}

\begin{abstract}
Previous Knowledge Distillation based efficient image retrieval methods employs a lightweight network as the student model for fast inference. However, the lightweight student model lacks adequate representation capacity for effective knowledge imitation during the most critical early training period, causing final performance degeneration. To tackle this issue, we propose a Capacity Dynamic Distillation framework, which constructs a student model with editable representation capacity. Specifically, the employed student model is initially a heavy model to fruitfully learn distilled knowledge in the early training epochs, and the student model is gradually compressed during the training.
To dynamically adjust the model capacity, our dynamic framework inserts a learnable convolutional layer within each residual block in the student model as the channel importance indicator. The indicator is optimized simultaneously by the image retrieval loss and the compression loss, and a retrieval-guided gradient resetting mechanism is proposed to release the gradient conflict. Extensive experiments show that our method has superior inference speed and accuracy, e.g., on the VeRi-776 dataset, given the ResNet101 as a teacher, our method saves 67.13\% model parameters and 65.67\% FLOPs without sacrificing accuracy.
\end{abstract}
\vspace{-4mm}

%%%%%%%%% BODY TEXT

\section{Introduction}
\label{sec:intro}

Image retrieval \cite{smoothap} aims to rank all the instances in a retrieval set based on their relevance to the query image. However, many image retrieval methods \cite{bcn,tits} use heavy networks to acquire a high accuracy, causing a low inference speed and hindering practical applications. As an efficient network compression technology \cite{kd, lasso_prune, dynamic}, knowledge distillation (KD) \cite{kd, dkr, dkd} has been widely validated to be useful for boosting the performance of the lightweight student model by transferring knowledge from a heavy teacher model, which is also applied to accelerate image retrieval, as done in \cite{cckd, vrkd, mbdl}.
% Image retrieval \cite{rdir, quad, csd}, i.e., given a query image, ranks all the instances in a retrieval set according to their relevance to the query. However, many image retrieval methods \cite{bcn,tits,quad} use heavy networks to acquire a high accuracy, causing a low inference speed and hindering practical applications of image retrieval. As an efficient network compression technology, knowledge distillation (KD) \cite{kd, dkr, dkd} has been widely validated to be useful for boosting the performance of the light weight model (student) by transferring knowledge from a heavy teacher network, which is also applied to accelerate image retrieval, as done in \cite{cckd, vrkd, mbdl}.

% , which widely used in various fields. For instance, person retrieval \cite{hybrid,bcn} and vehicle retrieval \cite{jqd, quad}, focuses on searching targets captured by different cameras installed at different locations of a city, which is great potential for smart cities.

%The mAP gain on ResNet50 significantly outperforms that of ResNet18 by 5.5\% mAP.

\begin{figure}[t]
	\centering
	\includegraphics[width=.90\linewidth]{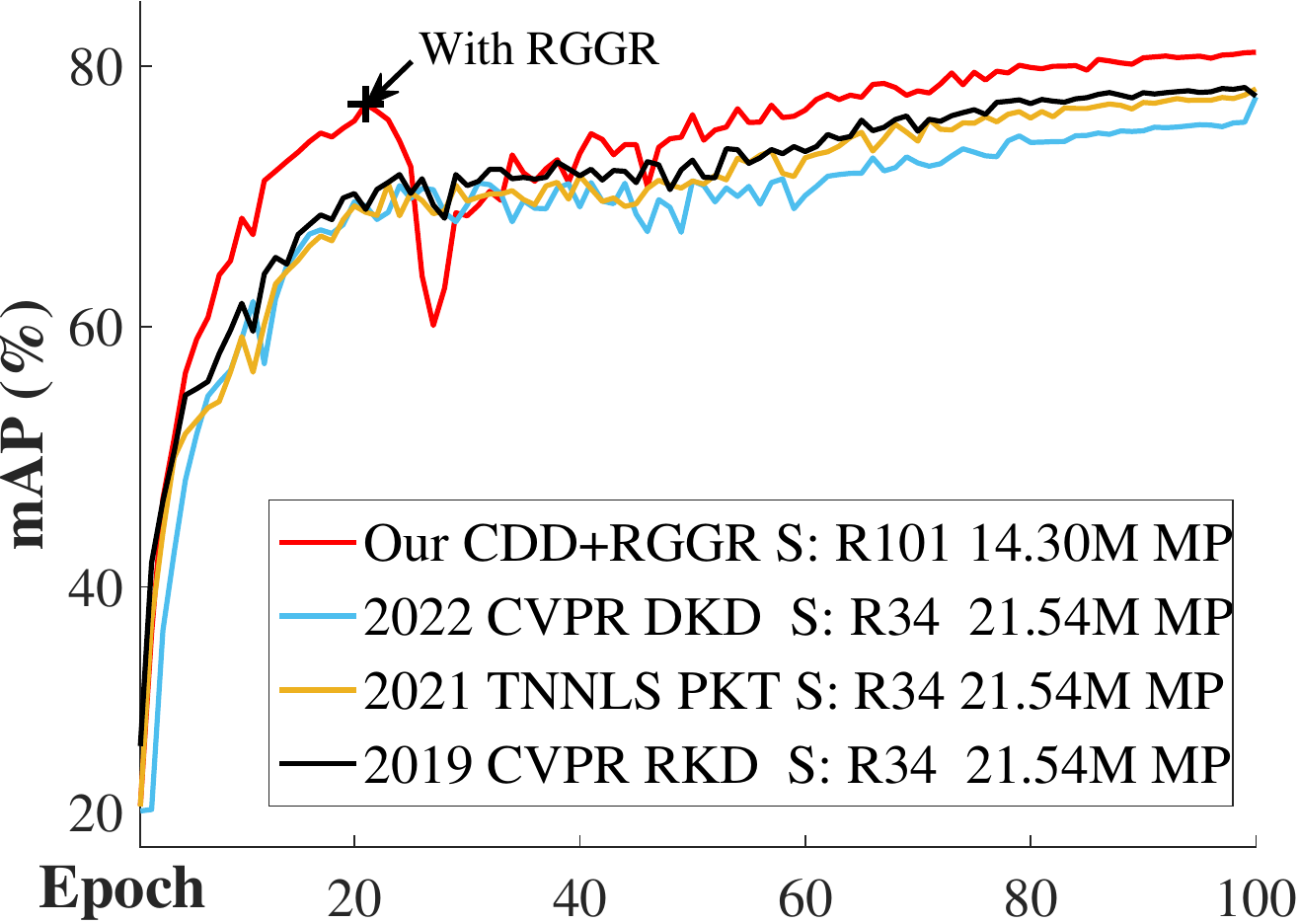}\vspace{-2mm}
	\caption{The mAP (\%) of KD methods in per epoch on VeRi776 \cite{veri}. R101 and R34 denote the ResNet101 and ResNet34 students, respectively. Our CDD+RGGR outperforms other KD methods because the heavy student has more talented in effectively learning distilled knowledge than a light student in the early epochs.}\label{fig:map}
	\vspace{-5mm}
\end{figure}

Previous KD-based image retrieval methods \cite{cckd, vrkd, rpkd} assign a lightweight network as the student model to acquire a fast inference speed.
However, we have an experimental observation that a lightweight student model is less talented in effectively learning distilled knowledge than a heavy student model in the early epochs, leading to final performance degeneration. As shown in Fig. \ref{fig:map}, a heavy ResNet101 (i.e., our) improves its performance faster than the light ones (ResNet34, other methods) in the early epochs (first 20 epochs) of KD optimization.
This finding can be compared to the human learning curve \cite{learning}, where a young kid can only comprehend a small portion of knowledge that is taught \cite{dkr}.
Many studies \cite{early, critical, earlycrop} have shown that critical network learning occurs during early training, determining the final solution of the associated optimization.
Furthermore, Cheng et al. \cite{discard} show that the network usually tries to model various visual concept knowledge in early epochs and then discard unimportant ones in later training.
%Speaking of KD in image retrieval, previous works \cite{cckd, vrkd, rpkd} usually assign a light network as the student network to acquire a friendly inference speed. However,  this introduces a challenge of light students only understanding a small portion of knowledge from the large teachers in early training because the representational capacity of students is much smaller than that of teachers as shown in Fig. \ref{fig:sketch}(a). {\color{red} This process is analogous to human learning curve \cite{learning} where a young kid can only comprehend a small portion of knowledge that is taught \cite{dkr}.} Furthermore, many studies \cite{early, critical, earlycrop, discard} have shown that critical aspects of artificial neural network learning occur during early training, determining the final solution of the associated optimization. For example, Cheng et al. \cite{discard} show that the neural network usually tries to model various visual concept knowledge in early training and then discard unimportant ones in later training. For that, the student's accuracy performance is limited if students only understand a small portion of teachers' knowledge in early training.

% \begin{figure}[tp]
% 	\centering
% 	\includegraphics[width=1.0\linewidth]{sketch.pdf}
% 	\caption{. }\label{fig:sketch}
% \end{figure}

Motivated by the above finding, we propose a new KD framework, \textbf{C}apacity \textbf{D}ynamic \textbf{D}istillation (CDD), to allow dynamic model compression during KD learning. Different from existing KD-based image retrieval methods that configure a lightweight student model, CDD employs a heavy initial network as the student model and thus the student has a high representation capacity for comprehensively understanding teachers' knowledge in the early KD iterations. To acquire a fast inference speed, we design a \textbf{D}istillation \textbf{G}uided \textbf{C}ompactor (DGC) module inserted after the heavy convolutional layer of the student as the channel importance indicator of each convolutional layer. Then, we implement a parametric sparse loss on DGC during KD learning to find the unimportant channel of the heavy convolutional layer, thus gradually reducing the capacity of the student network. The overall training process can be done end-to-end in one KD training period. After training, the sparse DGC will be pruned to a slim DGC, and the slim DGC and previous heavy convolutional layer can convert to a slim convolutional layer. Therefore, the heavy student model will be converted to a lightweight model.

To dynamically edit the student model capacity, DGC is optimized simultaneously by the image retrieval loss and the parametric sparse loss, resulting in a gradient conflict between the knowledge accumulation gradient generated from the image retrieval loss and the knowledge forgetting gradient generated from the parametric sparse loss. To release the gradient conflict, we propose a retrieval-guided gradient resetting mechanism (RGGR), which introduces a binary mask to zero out the knowledge accumulation gradient. Specifically, we first use the train data to simulate the retrieval result. Then, RGGR selects channels with little influence on simulation retrieval results and zeros the knowledge accumulation gradient, achieving a high prunability. As demonstrated in Fig.~\ref{fig:map}, when we activate RGGR (at 21-th epoch), the heavy student model focuses more on compression and suffers transient performance degradation. But, thanks to the well-trained KD optimization of students in the early epochs, the student model finally achieves a good balance between accuracy and inference speed.

The main contributions of this paper are listed as follows:

\begin{itemize}
       \vspace{-2mm}
		\item[(1)] We propose a capacity dynamic distillation framework (CDD) to effectively learn distilled knowledge in the early training epochs.
  \vspace{-1mm}
		\item[(2)] We propose retrieval-guided gradient resetting mechanism (RGGR) to release the gradient conflict between the image retrieval loss and the parametric sparse loss.
		\item[(3)] Extensive experiments demonstrate that our method is superior to state-of-the-art approaches in terms of inference and computations, by a large margin of 24.13\%MP and 21.94\% FLOPs.
\end{itemize}

\section{Related Works}\label{sec:rw}
%There are three key aspects highly related to this paper, namely, knowledge distillation (KD), network pruning (NP), and re-parameterization (Rep). In the following, we will comprehensively review those three aspects.

\tb{Network Pruning.}
Network Pruning (NP) \cite{lasso_prune, bn_prune,l2} aims to obtain a light network by removing unimportant parts from a well-trained yet large-scale network. Recent NP works primarily focus on structured pruning \cite{lasso_prune, bn_prune}, which applies sparsity functions to the convolutional layers of well-trained large models to filter out unimportant channels. However, NP often leads to a reduction in accuracy due to its irreversible sparsification process, which can cause significant damage to the network \cite{l2}.

\tb{Knowledge Distillation.}
Most KD-based image retrieval works \cite{cckd, mbdl, csd} employ a lightweight model as the student and transfers the knowledge among samples from teachers to guide the student optimization.
Although these works has helped light students improve accuracy, the weak representational capacity hinders further accuracy improvement of students. Therefore, recent KD works \cite{takd, dkr} have paid attention to this situation. However, they only focused on how to transfer knowledge well to facilitate students’ understanding of teachers' knowledge without considering enhancing students' representational capacity. 
Although self-distillation methods \cite{pskd, umts, mvkd}  assign a large student the same size as the teacher to better understand the teacher's knowledge, they suffer from large students' high computational cost in the inference phase. To address this issue, a natural solution is to aggregate NP with KD in a two-stage design. Specifically, the two-stage method \cite{cpk,kn,sar} first assign a large student to effectively distilled knowledge from teachers and then the well-trained large student network is pruned to a slim network. However, the two-stage approach may suffer a significant performance loss during the pruning stage since it still faces the challenge of sparsification process deviation. In this paper, we explore knowledge distillation with dynamic capacity compression, performed end-to-end, and the conflict between KD and NP is alleviated by a designed distillation guided compactor (DGC) module.

% For example, NPPM \cite{max} train an extra performance prediction network to predict the accuracy of sub-networks, which demonstrates the use of additional networks can improve the performance of NP.

%For the end-to-end way, NP and KD are simultaneously applied to a student has two advantages. (1) For KD, a heavy student network can be assigned to comprehensively understand teachers' knowledge in early training. Then, the heavy student can be pruned in the inference phase to acquire a high inference speed. (2) For NP, the sparse process of students can't severely deviate under the supervision of well-trained teachers. However, this end-to-end way exists a severe conflict between NP and KD because NP likes to forget knowledge but KD tends to learn more.

\tb{Re-parameterization.}
The re-parameterization (Rep) methods \cite{acnet, repvgg, DBB} construct a sequence of multiple convolutional layers to replace a single convolutional layer of an original network to enhance the feature learning ability of the original network in the training phase. Then, those sequences are simplified to a single convolutional layer to avoid extra computation consumption in the inference phase. Recent, Ding et al. \cite{resrep} proposed a gradient resetting and compactor re-parameterization (ResRep) method, which is the first attempt to apply Rep to NP. Motivated by this, we explore the gradient reseeting technique in knowledge distillation and propose zeroing out the selected channel's gradients according to feature retrieval results.
 %applies Rep to KD to address the problem that the light student has difficulty understanding teachers' knowledge. Besides, different from ResRep \cite{resrep} resetting gradient of the convolutional layer's channel with small weight value, our RGGR to zero out the convolutional layer's channel of according to feature retrieval results.

%  Technically speaking, Rep is not a model compressing method because it complicates a network during training and converts the complicated network into the original version during inference. But, the operation of converting sequences of multiple convolutional layers into a single convolutional layer has excellent potential for model compression. Most recently, Ding et al. \cite{resrep} proposed a gradient resetting and compactor re-parameterization (ResRep) method, which is the first attempt to applies Rep to NP. Furthermore, we attempt applies Rep to KD to address the problem that the light student has difficulty understanding teachers' knowledge.

\begin{figure}[tp]
    \centering
    \includegraphics[width=1.0\linewidth]{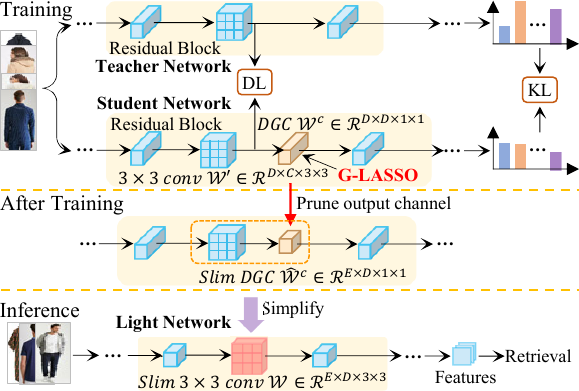}\vspace{-2mm}
    \caption{The capacity dynamic distillation framework. The high-capacity student can comprehensively understand teachers' knowledge in the early KD iterations. After training, the student can be converted into a light network to acquire a fast inference speed.}\label{fig:rc}
    \vspace{-4mm}
\end{figure}

\section{Method}\label{sec:app}
This section introduces the proposed CDD framework for efficient image retrieval. Unlike existing KD methods \cite{dkr, dkd} training a student model with fixed representation capacity, we present a new distillation pipeline for a dynamic representational capacity student model. The proposed distillation strategy employs a heavy student network to learn transferred knowledge effectively from teachers while gradually compressing the student’s capacity for obtaining a final model with fast inference speed. In our CCD framework, the DGC module is integrated to achieve dynamic representation compression during KD learning, as described in Section ~\ref{sec:31}.

Furthermore, we propose the RGGR mechanism to compress the student model by releasing the conflict between the retrieval-related losses and the compression loss. RGGR first evaluates the channel importance of students with respect to image retrieval. Then, the gradients derived from image retrieval-related losses are zeroed to eliminate the conflict in the final objective. This approach ensures that the weighting values of selected channels are solely optimized by the compression loss, enabling them to converge to zero more efficiently. The RGGR method will be discussed in detail in Section~\ref{sec:32}.

\subsection{Dynamic Representation Distillation}\label{sec:31}
The proposed CDD is illustrated in Fig. \ref{fig:rc}.
Similar to the typical KD methods, CCD is designed to transfer knowledge from a well-trained teacher model $P_{t}(y|x,\theta_{t})$ to an untrained student model $P_{s}(y|x,\theta_{s})$, where $x$ is the model input and $\theta$ denotes the model parameters.
Different from the previous methods employing models following the rule $|\theta_{s}|<|\theta_{t}|$ for model compression, CDD employs the student and the teacher satisfying $|\theta_{s}|\ge |\theta_{t}|$ for effective early learning, as discussed in Fig. ~\ref{fig:map}.

\subsubsection{Dynamic Representation Learning}

\tb{Distillation Learning.} From Fig. \ref{fig:rc}, we employ ResNet~ \cite{resnet} with cascaded residual blocks as an example.
During training, we consider globally and locally consistent distillation to transfer knowledge from teachers to students.

For global-consistent distillation, we use the Kullback-Leibler divergence (KL) loss $L_{kl}$ \cite{kd} to transfer logit knowledge, as done in
previous KD-based image retrieval works \cite{cckd, vrkd, mvkd}. For local-consistent distillation, different from layer-wise alignment adopted in previous methods, thanks to the large capacity student has the same network depth as teachers, we perform a block-wise alignment by focusing on the $3\times 3$ convolutional layers of each residual block to more completely transfers knowledge as follows:
\begin{equation}\label{eq:dl}
    L_{dl}(f^{t}, f^{s}) = \frac{1}{M}\sum \nolimits_{i=1}^{M}{\rm{DL}}\big({\rm{GAP}}(f^{t}), {\rm{GAP}}(f^{s})\big),
\end{equation}
where $f^{t}$ is the feature map output by the $3 \times 3$ convolutional layer of teachers and $f^{s}$ is the corresponding feature map from students; $M$ is the number of residual blocks, such as ResNet101 contains 33 residual blocks; $\rm{DL(\cdot,\cdot)}$ denotes Euclidean distance loss; The $\rm{GAP(\cdot)}$ denotes the global average pool layer to compress feature maps spatially.
%Notably, unlike previous KD works \cite{at,mvkd, mgd} that transfer feature knowledge in the residual layer, we can transfer feature knowledge at more subtle network structures (i.e., residual block). Because the residual block features channel of DRS is consistent with the teacher network, avoiding the harmful problem introduced by feature adaptive layers \cite{mgd}.}

\tb{Distillation Guided Compactor (DGC).}
To compress the student model dynamically during the training period, we add a proposed DGC module after each $3 \times 3$ convolutional layer of students.
Specifically, the DGC module here is a $1 \times 1$ convolutional layer without bias being initialized as an identity matrix, which is used to evaluate the importance of each channel of the $3 \times 3$ convolutional layer.
In this way, we can sparse the $1 \times 1$ convolutional layer during the training period. After training, the $3 \times 3$ convolutional layer can be pruned by merging these two layers. The merging operation is introduced in Section~\ref{sec:inference}.

To sparse the matrix, we implement G-LASSO loss $L_{np}$ \cite{group_lasso, resrep} on DGC to gradually decrease the student network's representational capacity as follows:
\begin{equation}\label{eq:np}
    L_{np}(\mathcal{W}^{c}) = \sum \nolimits_{i=1}^{D} \sqrt{\sum \nolimits_{j=1}^{D} (\mathcal{W}_{\scriptscriptstyle i,\scriptscriptstyle j, \scriptscriptstyle 1, \scriptscriptstyle 1}^{c})^2},
    \end{equation}
where $\mathcal{W}^{c}\in \mathbb{R}^{D \times D \times 1 \times 1}$ represents kernel parameters of DGC modules. The first and second dimensions of $\mathcal{W}^{c}$ represent the output and input channels, respectively.
%Also, the distillation constraint $L_{dl}$ is mainly learned by the $3 \times 3$ convolution with large representational capacity, and the sparsity constraint $L_{np}$  is only employed over the $1 \times 1$ convolution.

Finally, the student's overall loss $L_{all}$ includes $L_{dl}$ presented in Eq.~\ref{eq:dl}, KL loss $L_{kl}$, two widely used image retrieval losses and $L_{np}$ presented in Eq.~\ref{eq:np} as follows:
\begin{equation}\label{eq:total}
L_{all} = \frac{1}{2}L_{dl} + L_{id} + L_{tri} + L_{kl}  + \alpha L_{np},
\end{equation}
where $L_{id}$ is the label smooth regularization cross-entropy \cite{lsr}, as done in \cite{vrkd}; $L_{tri}$ is the triplet loss using hard sample mining strategy~\cite{triplet}; $\alpha$ is a hyper-parameter to control the sparsity level, with a default value of 0.004. 
% Please refer to the supplementary material for the influence of $\alpha$.

\noindent \tb{Discussion.} Intuitively, we argue that the DGC design is better than the previous two-stage ``self-KD+NP'' methods because the sparse process can't severely deviate under the supervision of the teacher network. Furthermore, the DGC design outperforms the previous end-to-end ``KD with NP'' methods because DGC can alleviate the conflict between KD and NP by decoupling their action positions.

\subsubsection{Efficient Inference with Pruning\label{sec:inference}}
After training, the output channel of DGC modules is treated as an unimportant channel and discarded if the output channel weight is less than the pruning threshold $\lambda$. The default value of $\lambda$ is $1 \times 10^{-5}$, as done in \cite{global, resrep}. Thus, the DGC module is trimmed to a slim one $\hat{\mathcal{W}}^{c}\in \mathbb{R}^{E \times D \times 1 \times 1}, E \le D$. Then, the $3 \times 3$ convolutional layer and the slim DGC module in one residual block can be simplified to being a slim $3 \times 3$ convolutional layer as follows:
\begin{equation}\label{eq:rc}
\mathcal{W} = T\big(T(\mathcal{W}') \circledast \mathcal{\hat{W}}^{c}\big), \quad B = B' \circledast \mathcal{\hat{W}}^{c},
\end{equation}
where $\mathcal{W} \in \mathbb{R}^{E \times C \times 3 \times 3} $ and $B \in \mathbb{R}^{E}$ represent the kernel parameters of the slim $3 \times 3$ convolutional layer and its bias, respectively.
$E$ and $C$ represent the output and input channel sizes of the slim $3 \times 3$ convolutional layer, respectively.
$T (\cdot)$ and $\circledast$ denote the transpose function and the convolution operator, respectively.
$\mathcal{W}' \in \mathbb{R}^{D \times C \times 3 \times 3} $ and $B' \in \mathbb{R}^{D}$ represent the kernel parameters of the $3 \times 3$ convolutional layers and its bias.

Finally, the $3 \times 3$ convolutional layers and the $\mathcal{\hat{W}}^{c}$ in each residual block are merged into a slim $3 \times 3$ convolutional layer. Besides, the last bottleneck layers (i.e., $1 \times 1$ convolutional layer) in each residual block also are thinned. For example, for the $1\times1$ convolutional layer after DGC, we can discard those unimportant channels according to the channel value in $\mathcal{\hat{W}}^{c}$. As a result, the heavy student model can be transformed into a lightweight network to acquire a fast inference speed for image retrieval.

%\noindent\textbf{Discussion}.
%
%
%
%Therefore, in later epochs, although the representational ability of DRS is already weaker than the constant weak representational student network in previous work \cite{mgd, dkd}, DRS still acquires a high accuracy performance because DRS obtain a well optimization in early epochs. Besides, DRKD can directly transfer intermediate features knowledge from teachers to DRS without any feature adaptive layer \cite{fitnets, ft, vid, crkd}

\begin{figure*}[tp]
	\centering
	\includegraphics[width=1\linewidth]{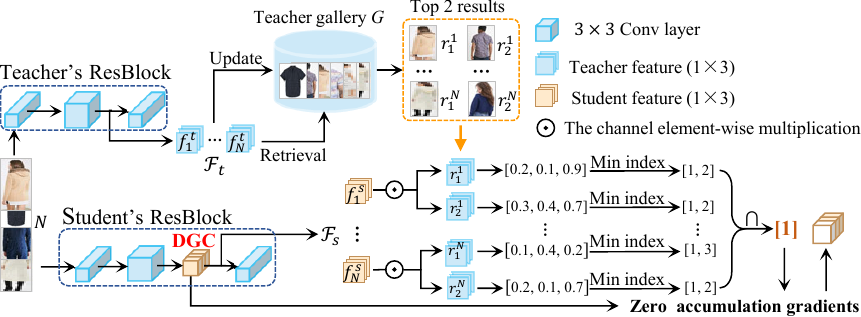}\vspace{-1mm}
	\caption{The retrieval-guided gradient resetting mechanism (RGGR). For ease of visualization, we assume that the DGC's output channel is 3. RGGR believes that the DGC's first channel has the most negligible influence on retrieval results by the absolute value of the channel product between $f_{s}^{i}$ and $r_{j}^{i}$. Then, RGGR resets the accumulation gradient of the first channel to 0 to push its weight value toward 0.}\label{fig:crgr}
    \vspace{-5mm}
\end{figure*}

\subsection{Retrieval-Guided Gradient Resetting}\label{sec:32}
Although CCD has achieved early learning with large representation capacity and efficient inference by dynamic compression, the optimization conflict remains unsolved. Specifically,  the gradient conflict between the knowledge accumulation gradient from image retrieval loss $L_{acc}=\frac{1}{2}L_{dl} + L_{id} + L_{tri} + L_{kl}$ and the knowledge forgetting gradient from pruning loss $L_{np}$ may cause a low compression rate.
More specifically, most of the output channels' weight of DGC may be updated to be approximated to 0 via $L_{np}$, but never close enough for perfect pruning due to $L_{acc}$~\cite{resrep}.
%Although DRD has been a good balance between inference speed and accuracy performance, it still has room for improvement in inference speed. Specifically, the gradient of DGC supervised by the G-LASSO loss function can be decoupled into the following two gradients \cite{resrep}: (1) The knowledge learning gradient generated by the network's accuracy performance-related objective function (e.g., triplet loss \cite{triplet} for image retrieval), which can improve the network's accuracy performance. (2) The knowledge forgetting gradient generated by the G-LASSO loss function to sparse the network without care for the accuracy degeneration. Thus, there are a gradient conflict between the knowledge learning gradient and the knowledge forgetting gradient, causing cannot achieve high prunability because most of the output channels' weight of DGC merely become closer to 0 than they used to be, but not close enough for perfect pruning \cite{resrep}.

To this end,  as shown in Fig. \ref{fig:crgr}, we propose the Retrieval-Guided Gradient Resetting mechanism (RGGR) to release this gradient conflict. Specifically,
RGGR employs a $D$-dimensional binary mask $M \in \{0, 1\}$ to zero the knowledge accumulation gradient of some DGC's output channels to acquire a resetting gradient $\hat{G}$ as follows:
\begin{equation}\label{eq:m_gradient}
\hat{G}(\mathcal{W}^{c}_i) = \frac{\partial L_{acc}(X, Y, \theta_s)}{\partial \mathcal{W}^{c}_i} M_i + \alpha \frac{\partial L_{np}(X, Y, \theta_s)}{\partial \mathcal{W}^{c}_i},
\end{equation}
where $\mathcal{W}^{c}_i= W^{c}_{i,:,:,:}$ denotes the $\mathcal{W}^c$ matrix's $i$-th output channel; $X$ and $Y$ are data examples and labels respectively.

From Eq. \eqref{eq:m_gradient},  with $M_{i} = 0$, the $i$-th output channel's knowledge accumulation gradient of DGC is reset to zero, causing the first term to be ignored. Thus, $M_i = 0$ can make DGC's $i$-th output channel weight value steadily move towards 0 to achieve high compression.
Intuitively, once the knowledge accumulation gradient of important output channels is zeroed, it would causes a decrease in the retrieval performance of students. Therefore, to maintain the student's retrieval performance, RGGR will zero out the knowledge learning gradient of the output channel that has the most negligible impact on the retrieval results as follows.

\tb{Retrieval Rank Matrix $R$ Formulation.} To evaluate the channel importance of each residual block with respect to image retrieval, we build a query set $\mathit{Q}$ and a gallery set $\mathit{G}$ on each block during training to simulate the retrieval results. Different from using only one batch of teacher features as the gallery set, we construct a large gallery set  $\mathit{G}$ to reveal the overall distribution of the data $X$ adequately.
Specifically, given a batch size of $N$ samples, we extract the teacher feature $\mathcal{F}_t = [f_{\scriptscriptstyle 1}^{t}, f_{\scriptscriptstyle 2}^{t},..., f_{\scriptscriptstyle N}^{t}] \in \mathbb{R}^{N \times D}$ from the $3 \times 3$ convolutional layer and GAP layers to update the gallery $\mathit{G}$.
The gallery set is implemented in a queue with a fixed size $L$, which means the first-in-first-out policy is followed to maintain the queue length, as done in \cite{queue, crkd}.
%Fortunately, our framework does not need to optimize the gallery $\mathit{G}$. Hence, the use of the gallery adds a small amount of video memory overhead.
% It is done by maintaining an instance queue of size $L$ in each residual block for storing output features from the teacher model, as done in \cite{queue, crkd}.

For building the query set $\mathit{Q}$, we should extract the student feature $\mathcal{F}_s = [f_{\scriptscriptstyle 1}^{s}, f_{\scriptscriptstyle 2}^{s},..., f_{\scriptscriptstyle N}^{s}] \in \mathbb{R}^{N \times D}$ from DGC and GAP layers as the query set $\mathit{Q}$. However, $\mathcal{F}_s$ is not invariant and semantic enough for presenting the image information during early training, causing the matching results between $\mathcal{F}_s$ and $\mathcal{F}_t$ to be unreliable. Therefore, we use $\mathcal{F}_t$ instead of $\mathcal{F}_s$ as the query set to retrieve the gallery set $\mathit{G}$.
Formally, we obtain the retrieval rank matrix $R \in \mathbb{R}^{N \times L}$ as follows:
\begin{equation}\label{eq:r}
R = \{r_{\scriptscriptstyle j}^{i} \in \mathbb{R}^{1 \times D} \big| 1\le i \le N, 1\le j \le L \},
\end{equation}
where $r_j^{i}$ represents the $i$-th query feature retrieved from the gallery set $G$ and then returning the $j$-th gallery feature.

\tb{Binary Mask $M$ Formulation based on $R$.}
Since the retrieval rank matrix $R$ is obtained by sorting the retrieval distance matrix, we conclude that $R$ can indicate the output channels with the most negligible impact on the retrieval result.
In addition, to reduce the negative impact of gallery features with low relevance, we select top-K retrieval gallery feature from $R$ as the retrieval result $R_{\scriptscriptstyle K} \in \mathbb{R}^{N \times K}$ to find unimportant output channel as follows:
\begin{equation}\label{eq:rk}
R_{\scriptscriptstyle K} = \{r_{\scriptscriptstyle j}^{i} \in \mathbb{R}^{1 \times D} \big| 1\le i \le N, 1\le j \le K\},
\end{equation}
where $K=2$ empirically.

Thus, the unimportant output channel index $I$ between the query set $Q$ and the retrieval result $R_K$ is as follows:
\begin{equation}\label{eq:total_index}
I = \mathop{\cap}_{i=1}^{N} \mathop{\cap}_{j=1}^{K} I_{ij},
\end{equation}
where $I_{ij}$ denotes the unimportant output channel index between the $i$-th query feature $f_{\scriptscriptstyle i}^{s}$ and $r_{\scriptscriptstyle j}^{i}$ as follows:
\begin{equation}\label{eq:index}
I_{ij} = \omega\big(A (f_{\scriptscriptstyle i}^{s},r_{\scriptscriptstyle j}^{i})\big), 1\le i \le N, 1\le j \le K,
\end{equation}
where $ \omega(\cdot)$ represents a series of operations, including sorting the importance of the output channels in ascending order, picking the smallest one from the ordered index queue, storing its index, and stopping picking when $p=50\%$ ratios of the index have been selected.
The value of $p$ is chosen empirically. The $A (\cdot)$ is a function that evaluates the importance of DGC modules' each output channel based on the absolute value of the product of $f_{s}^{i}$ and $r_{j}^{i}$ as follows:

% aking the typical cosine metric \cite{hybrid, bcn} as an example, $\mathcal{D}_{imp}(\cdot)$ should reveal the degree of relevance between $f_{s}^{i}$ and $r_{j}^{i}$ in the space of cosine metric $A$:
% \begin{equation}\label{eq:cosine_dist}
% A = \frac{\sum\nolimits_{d=1}^{D} f_{i, d}^{s} r_{j, d}^{i}}{\Vert f_{s}^{i} \Vert \Vert r_{j}^{i} \Vert}, 1 \le  i \le N, 1 \le j \le K,
% \end{equation}
% where $\Vert \cdot \Vert$ represents the L2-norm.

% From Eq. \eqref{eq:cosine_dist}, we can find that the absolute value of the product of the corresponding channel of the two features can measure the importance of the channel. Thus, the $\mathcal{D}_{imp}$ is formulated as follows:
\begin{equation}\label{eq:imp}
A (f_{\scriptscriptstyle i}^{s},r_{\scriptscriptstyle j}^{i})= \big[A_d \big],  A_d= \big|f_{i, d}^{s}\cdot r_{j, d}^{i} \big|,  1 \le d \le D,
\end{equation}
where $f_{i, d}^{s}$  and $r_{j, d}^{i}$ represent the $d$-th channel value of $f_{s}^{i}$ and the $d$-th channel value of $r_{j, d}^{i}$, respectively.

Overall, based on the unimportant output channel index $I$, we update the binary mask $M$ to zero the knowledge accumulation gradient of the unimportant output channel of DGC as follows:
\begin{equation}\label{eq:reset_mask}
M = [M_{i}=0 \big|i \in I,1 \le i \le D].
\end{equation}
% The more algorithm details of Binary Mask $M$ generation please refer to the supplementary material.

	\begin{table*}[tp]
		\caption{Ablation results on In-Shop \cite{inshop} and VeRi776 \cite{veri}.\vspace{-3mm}}\label{tab:as}
		\newcommand{\tabincell}[2]{\begin{tabular}{@{}#1@{}}#2\end{tabular}}
		\renewcommand\arraystretch{1}
		\centering
		\setlength{\tabcolsep}{3.5pt}
		\begin{threeparttable}
			\resizebox{1.0\textwidth}{!}{
				\begin{tabular}{cccccccccc  cccc cccc }
					\hlinewd{1.5pt}
					\multirow{2}{*}{METHOD}
					&\multirow{2}{*}{TEACHER}
					&\multirow{2}{*}{STUDENT}
					&\multicolumn{4}{c}{In-Shop}
					& \multicolumn{4}{c}{VeRi776}\\
					\cline{4-11}
					%&KD
					&&&MP (M)& FLOPs (G)  &mAP(\%)     &R1(\%) &MP (M)& FLOPs (G)  &mAP(\%)     &R1(\%)\\
					\hlinewd{0.8pt}
					Teacher & - &ResNet101    &43.50 &12.99 &\textbf{81.72} &\textbf{95.23} &43.50 &12.99 &80.50 &96.42\\	
					CDD w/o DGC &ResNet101 &ResNet101   &35.89 &10.94  &71.87 &90.45 &39.80 &11.96  &62.14 &93.27\\
					CDD &ResNet101 &ResNet101  &20.14	&6.31 &81.38	&95.01  &18.59	&5.77 &80.36	&96.48\\
					CDD+RGGR  &ResNet101 &ResNet101 &\textbf{14.97} 	&\textbf{4.68}  &81.28	&95.14 &\textbf{14.30}	&\textbf{4.46}  &\textbf{80.67}	&\textbf{96.66}\\
					\hlinewd{1.5pt}
			\end{tabular}}
		\end{threeparttable}
	\end{table*}

\section{Experiments}\label{sec:exp}
    To demonstrate the superiority of our CDD+RGGR method, we conduct evaluations on three widely used public image retrieval datasets: In-Shop \cite{inshop}, VeRi776 \cite{veri}, and MSMT17 \cite{msmt17}. In what follows, we briefly introduce datasets and performance metrics. Then, we conduct ablation experiments and compare CDD+RGGR with state-of-the-art methods. we conduct analysis experiments to examine the role of CDD+RGGR comprehensively.
    
    % hyper-parameter analysis experiments to comprehensively examine the role of RGGR and present visual results that highlight the influence of RGGR on DGC's output channel number.

\subsection{Datasets and Performance Metric}
   \textbf{In-Shop Clothes Retrieval (In-Shop)} \cite{inshop} is a commonly used clothes retrieval database, which contains 72,712 images of clothing items belonging to 7,986 categories. The training set includes 3,997 classes with 25,882 images. The query set with 14,218 images of 3,985 classes. The gallery set has 3,985 classes with 12,612 images.

   \textbf{VeRi776} \cite{veri} is a vehicle retrieval dataset. The training set contains 37,746 images of 576 subjects. The query set of 1,678 images of 200 subjects. The gallery set of 11,579 images of the same 200 subjects.
%constructed by 20 cameras in unconstrained traffic scenarios

   \textbf{MSMT17} \cite{msmt17} is the largest pedestrian retrieval database, which contains 126,441 images of 4,101 pedestrian identities. The training set includes 32,621 training images of 1,041 identities. The test set includes 11,659 query images and 82,161 gallery images of 3,060 identities.
%captured by 3 indoor cameras and 12 outdoor cameras

   The \textbf{Cosine} distance between features as the retrieval algorithm, i.e., the more similar the gallery image, the higher the ranking. The mean average precision (mAP) \cite{duke} and rank-1 identification rate (R1) \cite{veri} both are applied to evaluating the retrieval accuracy performance. The number of model parameters (MP) and floating-point of operations (FLOPs) are used to measure the model size, the computational complexity, respectively.

%and the feature extraction time (FET) per image \cite{hybrid}

\subsection{Implementation Details}
   The software tools are Pytorch 1.12 \cite{pytorch}, CUDA 11.6, and python 3.9. The hardware device is one GeForce RTX 3090Ti GPU. The network training configurations are as follows.
   (1) We use ImageNet pre-trained ResNet101 \cite{resnet} as backbones and set the last stride of ResNet101 to 1, as done in \cite{vrkd, mbdl}.
   (2) The teacher network is frozen during students' training.
   (3) The data augmentation includes z-score normalization, random cropping, random erasing \cite{earse}, and random horizontal flip operations, as done in \cite{vrkd}.
   (4) The mini-batch stochastic gradient descent method \cite{alexnet} is used as the optimizer. The mini-batch size is set to 96, including 16 identities, and each identity holds 6 images.
   (5) Setting weight decays as $5\times10^{-4}$ and momentums as 0.9.
   (6) The cosine annealing strategy \cite{cosine_anneal} and linearly warmed strategy \cite{warmup} are applied to adjust learning rates.
   (7) The learning rates are initialized to $1 \times 10^{-3}$, then linearly warmed up to $1 \times 10^{-2}$ in the first 10 epochs, the drop point for learning rates is the 40-th epoch. For MSMT17 \cite{msmt17}, the total training epoch is 120. For In-Shop \cite{inshop} and VeRi776 \cite{veri}, the total training epoch is 100.
   (8) Considering that pedestrians datasets and other datasets have different different aspect ratios, for MSMT17 \cite{msmt17}, the image resolution is set as $320\times 160$. For In-Shop \cite{inshop} and VeRi776 \cite{veri}, the image resolution is set as $256\times256$.

%-------------------------------------------------------------------------

\subsection{Ablation Experiments}
 As shown in Table \ref{tab:as}, we conduct ablation experiments on In-Shop \cite{inshop} and VeRi776 \cite{veri}. Teacher means that we directly use the ResNet101 teacher to evaluate performance. CDD w/o DGC means that the student discards DGC module, causing KD of Eq. \ref{eq:dl} and NP of Eq. \ref{eq:np} to act on the same $3 \times 3$ convolutional layers.

First, compared to Teacher, we can find that CDD w/o DGC slightly wins MP and FLOPs but severely degrades R1 and mAP. Specifically, CDD w/o DGC is inferior to Teacher by 18.36\% mAP and 3.15\% R1 on VeRi776 \cite{veri}. These results show that playing KD and NP on the same convolutional layers can compress networks but is injured for retrieval performance. The fault lies in the potential conflict between KD and NP because KD guides students to inherit more teachers' knowledge, but NP encourages students to forget knowledge. Furthermore, with the usage of DGC, CDD outperforms CDD w/o DGC in terms of all metrics. For example, on VeRi776 \cite{veri}, CDD outperforms CDD w/o DGC by 18.22\% mAP, 3.21\% R1, 21.21M MP, and 6.19G FLOPs. These comparisons demonstrate that DGC can alleviate the conflict between KD and NP.

Second, with using RGGR, the student further improved inference performance without loss of accuracy performance. Compared with CDD, CDD+RGGR saves 25.67\% MP and 25.80\% FLOPs on In-Shop \cite{inshop}, and 23.08\% MP and 22.70\% FLOPs on VeRi776 \cite{veri}. The comparison demonstrates that RGGR can further boost the inference performance of students without sacrificing accuracy.

\begin{table}[tp]
	\caption{Performance comparison on In-Shop.\vspace{-2mm} }.\label{tab:sota_inhop}
	\newcommand{\tabincell}[2]{\begin{tabular}{@{}#1@{}}#2\end{tabular}}
	\renewcommand\arraystretch{1.0}
	\centering
	\setlength{\tabcolsep}{1pt}
	%	\footnotesize
	\small
	\begin{threeparttable}
		\scalebox{0.96}{
			\setlength{\tabcolsep}{1pt}
			\begin{tabular}{c  c c cccc }
				\hlinewd{1.5pt}
				METHOD \tnote{1}
				&TEACHER
				&STUDENT
				& \begin{tabular}{c}
					MP\\
				\end{tabular}
				& \begin{tabular}{c}
					FLOPs\\
				\end{tabular}
				&\begin{tabular}{c}
					mAP\\
				\end{tabular}
				& \begin{tabular}{c}
					R1\\
				\end{tabular} \\
				\hlinewd{0.8pt}
				Fastretri \cite{fastreid}&\multirow{1}{*}{{-}}&ResNet50 &25.54 &8.13  &-  &91.97\\
				\hlinewd{0.8pt}		
				%KD \cite{kd} $\ast$ &ResNet101& ResNet34 &21.54&7.31 &80.56	&94.66   \\
				CCKD \cite{cckd} $\ast$  &ResNet101& ResNet34 &21.54 &7.31 &80.65 &94.78\\
				  VID \cite{vid} $\ast$  &ResNet101& ResNet34 &21.54 &7.31 &80.61 &94.57 \\
				FT \cite{ft} $\ast$ &ResNet101& ResNet34 &21.54&7.31  &80.51 &94.55 \\

				SP \cite{spkd} $\ast$ &ResNet101& ResNet34 &21.54&7.31  &80.81 &94.89 \\
				AT \cite{at} $\ast$ &ResNet101& ResNet34 &21.54&7.31  &80.85	&94.91 \\
				RKD \cite{rkd} $\ast$ &ResNet101& ResNet34 &21.54 &7.31 &81.03 &94.81 \\
				PKT \cite{pkt} $\ast$  &ResNet101& ResNet34 &21.54 &7.31 &80.59 &94.68 \\
				MBDL \cite{mbdl} $\ast$  &ResNet101& ResNet34 &21.54&7.31  &80.76 &94.71\\
				DKD \cite{dkd} $\ast$  &ResNet101& ResNet34 &21.54 &7.31 &80.93 &94.86\\
                CSKD \cite{csd} $\ast$  &ResNet101& ResNet34 &21.54 &7.31 &80.65 &94.77\\
                KDPE \cite{kdpe} $\ast$  &ResNet101& ResNet34 &21.54 &7.31 &78.99 &94.08\\
				ResRep \cite{resrep}$\ast$  &ResNet101& ResNet101 &16.27  &5.11  & 79.86  &94.68\\
				CDD+RGGR   &ResNet101& ResNet101 &\textbf{14.97} & \textbf{4.68}   & \textbf{81.28} & \textbf{95.14} \\
				\hlinewd{1.5pt}
			\end{tabular}
		}
		\begin{tablenotes}
			\footnotesize
			\item[1]{The $\ast$ represents the result is re-implemented.}
		\end{tablenotes}
	\end{threeparttable}
	\vspace{-2mm}
\end{table}

\begin{table}[tp]
\caption{Performance comparison on VeRi776.\vspace{-2mm}}\label{tab:sota_veri776}
\newcommand{\tabincell}[2]{\begin{tabular}{@{}#1@{}}#2\end{tabular}}
\renewcommand\arraystretch{1.0}
\centering
\setlength{\tabcolsep}{3.5pt}
%	\footnotesize
\small
\begin{threeparttable}
	\scalebox{0.96}{
		\setlength{\tabcolsep}{1pt}
		\begin{tabular}{c cc cccc }
			\hlinewd{1.5pt}
			METHOD \tnote{1}
			&TEACHER
			&STUDENT
			&MP& FLOPs  &mAP     &R1\\
			\hlinewd{0.8pt}
			PVEN \cite{pven} &\multirow{2}{*}{{-}} &ResNet50  &25.54&8.13 &79.50 &95.60 \\
			ViT-Base \cite{vit} $\ast$  & &ViT \cite{vit} &86.7& 55.6 &78.92 &95.84\\
			%GiT-Base \cite{git} $\ast$ &\XSolidBrush&\XSolidBrush&  &ViT \cite{vit}  &92.1 &55.7 &80.34 &96.86\\
			\hlinewd{0.8pt}	
			%KD \cite{kd} $\ast$  &ResNet101 & ResNet34  &21.54&7.31  &76.42	 &95.47\\
			%KD-FitNet \cite{fitnets} $\ast$  &R101& R34    &21.54&7.31  &76.87&95.41\\
			CCKD \cite{cckd}$\ast$   &ResNet101 & ResNet34  &21.54 &7.31 &76.34 &95.23\\
			VID \cite{vid} $\ast$  &ResNet101 & ResNet34 &21.54 &7.31 & 76.92    &95.35\\
			FT \cite{ft} $\ast$   &ResNet101 & ResNet34 &21.54&7.31  &77.18 &95.77\\
			AT \cite{at} $\ast$   &ResNet101 & ResNet34 &21.54&7.31  &77.98  &95.41\\
			SP \cite{spkd} $\ast$  &ResNet101 & ResNet34 &21.54&7.31  &77.50 &95.77  \\
			RKD \cite{rkd} $\ast$   &ResNet101 & ResNet34 &21.54&7.31 &78.56 &95.65\\
			PKT \cite{pkt} $\ast$   &ResNet101 & ResNet34 &21.54 &7.31 &78.45  & 95.47\\
			MBDL \cite{mbdl} $\ast$   &ResNet101 & ResNet34   &21.54&7.31  & 77.91&95.53\\
			DKD  \cite{dkd} $\ast$   &ResNet101 & ResNet34   &21.54&7.31 &76.40 &95.59\\
            CSKD \cite{csd} $\ast$   &ResNet101 & ResNet34 &21.54 &7.31 &73.60 &94.34\\
            KDPE \cite{kdpe} $\ast$   &ResNet101 & ResNet34 &21.54 &7.31 &74.27 &94.93\\
			ResRep \cite{resrep} $\ast$   &ResNet101 & ResNet101 & 15.87 &4.87   &77.60  &95.11 \\
			CDD+RGGR& ResNet101 & ResNet101 &\textbf{14.30}	&\textbf{4.46}  &\textbf{80.67} &\textbf{96.66}\\
			
			\hlinewd{0.8pt}
			UMTS \cite{umts} &ResNet50  & ResNet50   &25.54&8.13  &75.9& 95.8\\
			VKD \cite{mvkd}  &ResNet50   & ResNet50  &25.54&8.13  &\textbf{79.17}&95.23\\
			ResRep \cite{resrep} $\ast$ &ResNet50 & ResNet50 &11.24 &3.83 &75.74 &94.66 \\
			CDD+RGGR &ResNet50  &ResNet50 &\textbf{9.33}&\textbf{3.20} &78.75 &\textbf{95.95}\\

			\hlinewd{1.5pt}
		\end{tabular}
		  }
	\begin{tablenotes}
		\footnotesize
		\item[1]{The $\ast$ represents the result is re-implemented.}
	\end{tablenotes}
\end{threeparttable}
\end{table}

\subsection{Comparison with State-of-the-art Methods}
Table \ref{tab:sota_inhop}, \ref{tab:sota_veri776} and \ref{tab:sota_msmt17} summarize comparisons on In-Shop \cite{inshop},  VeRi-776 \cite{veri}, and  MSMT17 \cite{msmt17} datasets, respectively. Here, both MP (M), FLOPs (G), mAP (\%), R1 (\%) are obtained by only using the student during inference. For a fair comparison, Hinton’s original Kullback-Leibler divergence strategy \cite{kd} is applied to all compared re-implemented KD methods, as done in previous works \cite{cckd, mybmvc, mvkd}. 
Moreover, an advanced NP method (i.e., ResRep \cite{resrep}) was also re-implemented to compare with our method. The comparison analyses are discussed as follows.
% Besides, the backbone network is evaluated to estimate the inference performance of non-compressed methods.

\begin{table}[tp]
	\caption{Performance comparison on MSMT17.\vspace{-2mm}}\label{tab:sota_msmt17}
	\newcommand{\tabincell}[2]{\begin{tabular}{@{}#1@{}}#2\end{tabular}}
	\renewcommand\arraystretch{1}
	\centering
	\setlength{\tabcolsep}{3.5pt}
	%	\footnotesize
	\small
	\begin{threeparttable}
	\scalebox{1}{
		\setlength{\tabcolsep}{1pt}
		\begin{tabular}{c  c c cccc c}
			\hlinewd{1.5pt}
			METHOD \tnote{1}
			&TEACHER
			&STUDENT
			&MP& FLOPs  &mAP    & R1\\
			\hlinewd{0.8pt}
			IANet \cite{ianet} &\multirow{2}{*}{{-}}  &ResNet50 &25.54 &4.07  &46.8  &75.5\\
			BINet \cite{binet}   &  &ResNet50 & 25.54 &4.07 & 52.8 & 76.1 \\
			
			\hlinewd{0.8pt}
            CCKD \cite{cckd}$\ast$  &ResNet101 & ResNet34  &21.54 &5.71 &56.65 &80.58\\
            VID \cite{vid} $\ast$  &ResNet101 & ResNet34 &21.54 &5.71 &56.59 &80.06 \\
            FT \cite{ft} $\ast$  &ResNet101 & ResNet34 &21.54&5.71  &56.67 &80.38 \\
            AT \cite{at} $\ast$  &ResNet101 & ResNet34 &21.54&5.71  &58.70 &81.62 \\
            SP \cite{spkd} $\ast$  &ResNet101 & ResNet34 &21.54&5.71  &57.32 &80.37 \\
            RKD \cite{rkd} $\ast$  &ResNet101 & ResNet34 &21.54&5.71 &57.97 &81.02\\
            PKT \cite{pkt} $\ast$  &ResNet101 & ResNet34 &21.54 &5.71 &57.02 &80.31\\
			MBDL \cite{mbdl} $\ast$ &ResNet101 & ResNet34   &21.54&5.71 &57.18 &80.50\\
            DKD \cite{dkd} $\ast$ &ResNet101 & ResNet34  &21.54&5.71 &56.95 &80.63\\
            CSKD \cite{csd} $\ast$  &ResNet101 & ResNet34 &21.54 &7.31 &54.89 &79.17\\
            KDPE \cite{kdpe} $\ast$  &ResNet101 & ResNet34 &21.54 &7.31 &52.45 &78.18\\
            ResRep \cite{resrep} $\ast$ &ResNet101 &ResNet101 &16.49 &3.94 &52.69 &75.63\\
            CDD+RGGR  &ResNet101  &ResNet101 &\textbf{15.00} &\textbf{3.66} &\textbf{60.98} &\textbf{81.68}   \\
			\hlinewd{1.5pt}
		\end{tabular}
	}
	\begin{tablenotes}
		\footnotesize
		\item[1]{The $\ast$ represents the result is re-implemented.}
	\end{tablenotes}
	\end{threeparttable}
%	\vspace{-.5cm}
\end{table}

\tb{In-Shop dataset.}
Table \ref{tab:sota_inhop} shows that CDD+RGGR has several significant advantages over other compression methods. Firstly, when using the same ResNet101 teacher, CDD+RGGR achieves the highest performance across all metrics compared to other KD methods. Specifically, regarding inference performance, CDD+RGGR outperforms these methods by 5.27M MP and 2.63G FLOPs. Secondly, compared to ResRep \cite{resrep}, CDD+RGGR achieves a higher accuracy performance of 1.42\% mAP and 0.46\% R1 while maintaining similar inference performance. 
%Overall, CDD+RGGR is state-of-the-art on the In-Shop dataset \cite{inshop}.

  \begin{figure*}[tp]
	\centering
	\includegraphics[width=1\linewidth]{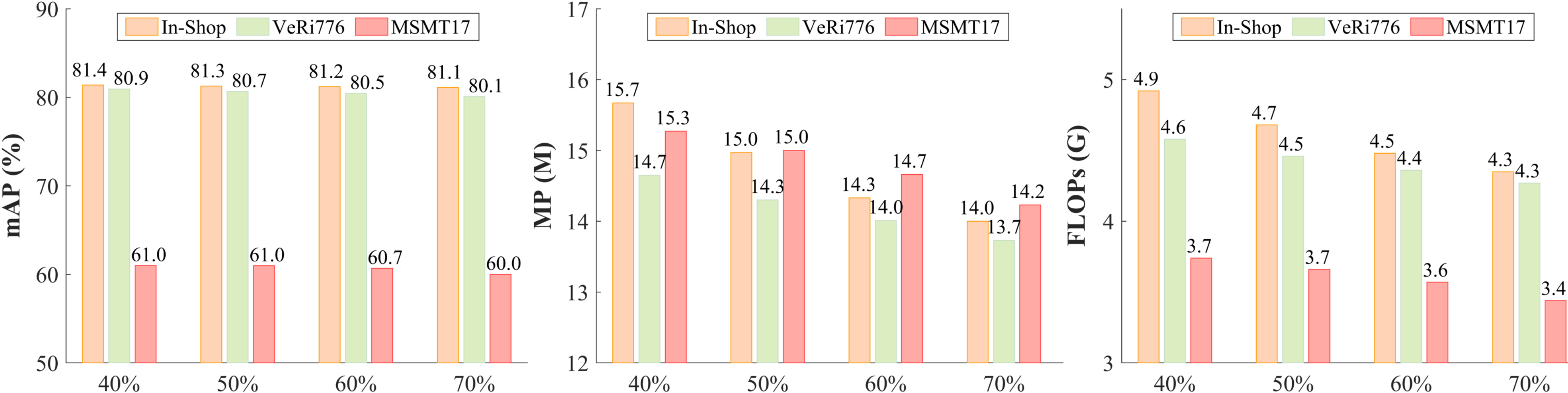}\vspace{-1mm}
	\caption{The influences of $p$ values. (a) on mAP. (b) on MP. (c) on FLOPs. As $p$ increases, the knowledge accumulation gradient of more channels is zeroed, resulting in a slight decrease in mAP and a noticeable improvement in inference performance.}\label{fig:p}
	\vspace{-2mm}
\end{figure*}

\begin{figure}[tp]
	\centering
	\includegraphics[width=0.8\linewidth]{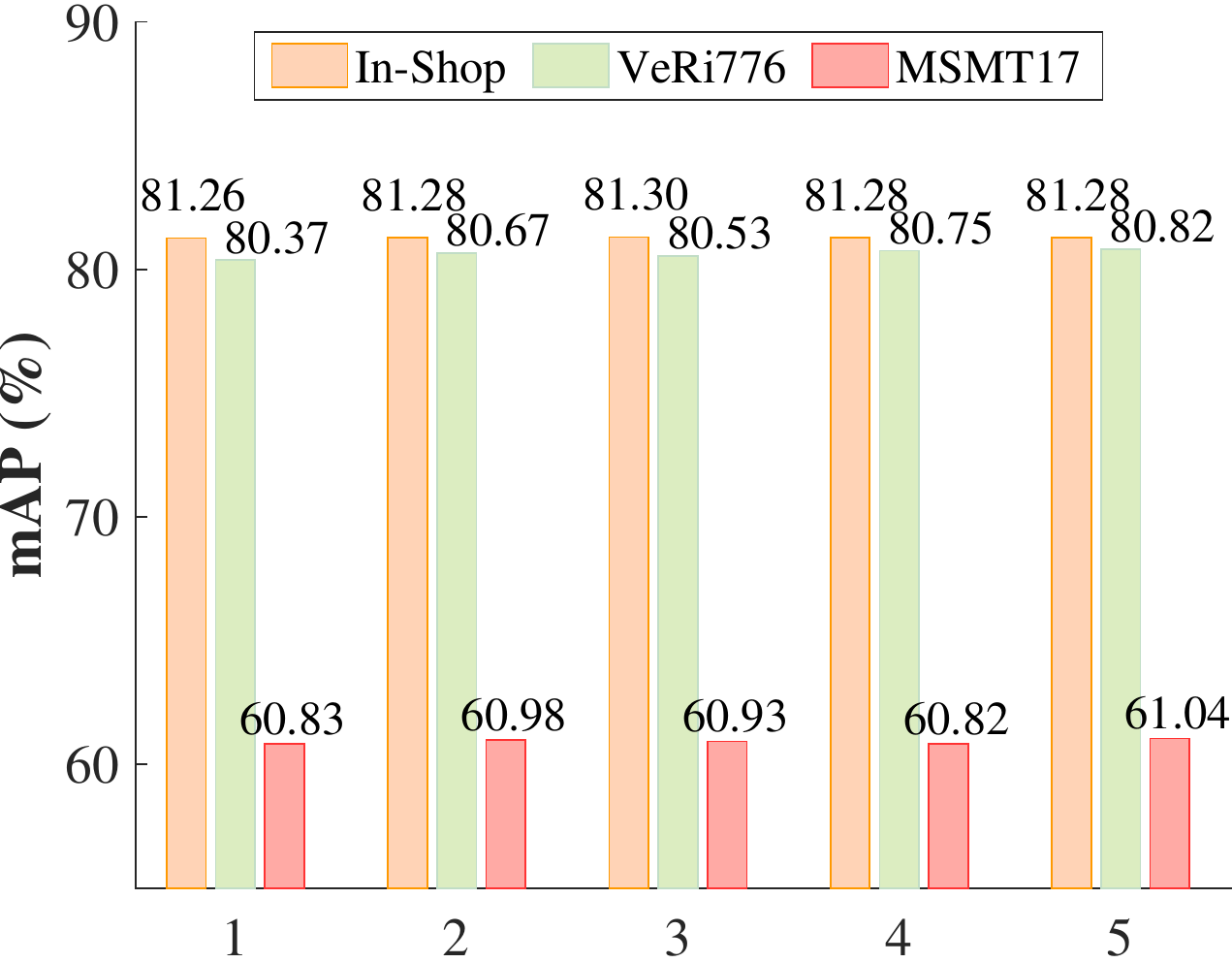} \vspace{-1mm}
	\caption{The $K$ influence on mAP.}\label{fig:K}
	\vspace{-2mm}
\end{figure}

\tb{VeRi776 dataset.}
   From Table  \ref{tab:sota_veri776}, the compressed method CDD+RGGR outperforms non-compressed methods like PVEN \cite{pven} and ViT \cite{vit}, which use complex backbone networks. Compared with other compressed methods that use the same ResNet101 teacher, CDD+RGGR can achieve higher accuracy with a lighter and more efficient student model with 14.30M MP and 4.46G FLOPs, along with 80.67\% mAP and 96.66\% R1. Specifically, regarding mAP accuracy, CDD+RGGR outperforms RKD \cite{rkd} by 2.11\%, PKT \cite{pkt} by 2.22\%, and all other competitors by a significant margin while using only 61.01\% FLOPs of other KD methods.
   Besides, using ResNet50 as the teacher, CDD+RGGR loses first place in mAP to VKD \cite{mvkd}, a self-distillation method that focuses on improving the accuracy of the original model (i.e., teacher models) itself rather than educating a lightweight student. Nevertheless, our approach is still valuable because CDD+RGGR can save 63.47\% MP and 59.33\% FLOPs with a slight loss in accuracy.

    %These mAP and R1 comparisons may seem unfair due to student networks are not consistent, but please note that our ERHC are compressed to be the smallest one. Our ERHC aims to learn a lightweight student and to maintain accuracy performance. We think it's acceptable to have a significant reduction in computations with a tiny loss of mAP.

	\begin{figure}[tp]
		\centering
		\includegraphics[width=0.8 \linewidth]{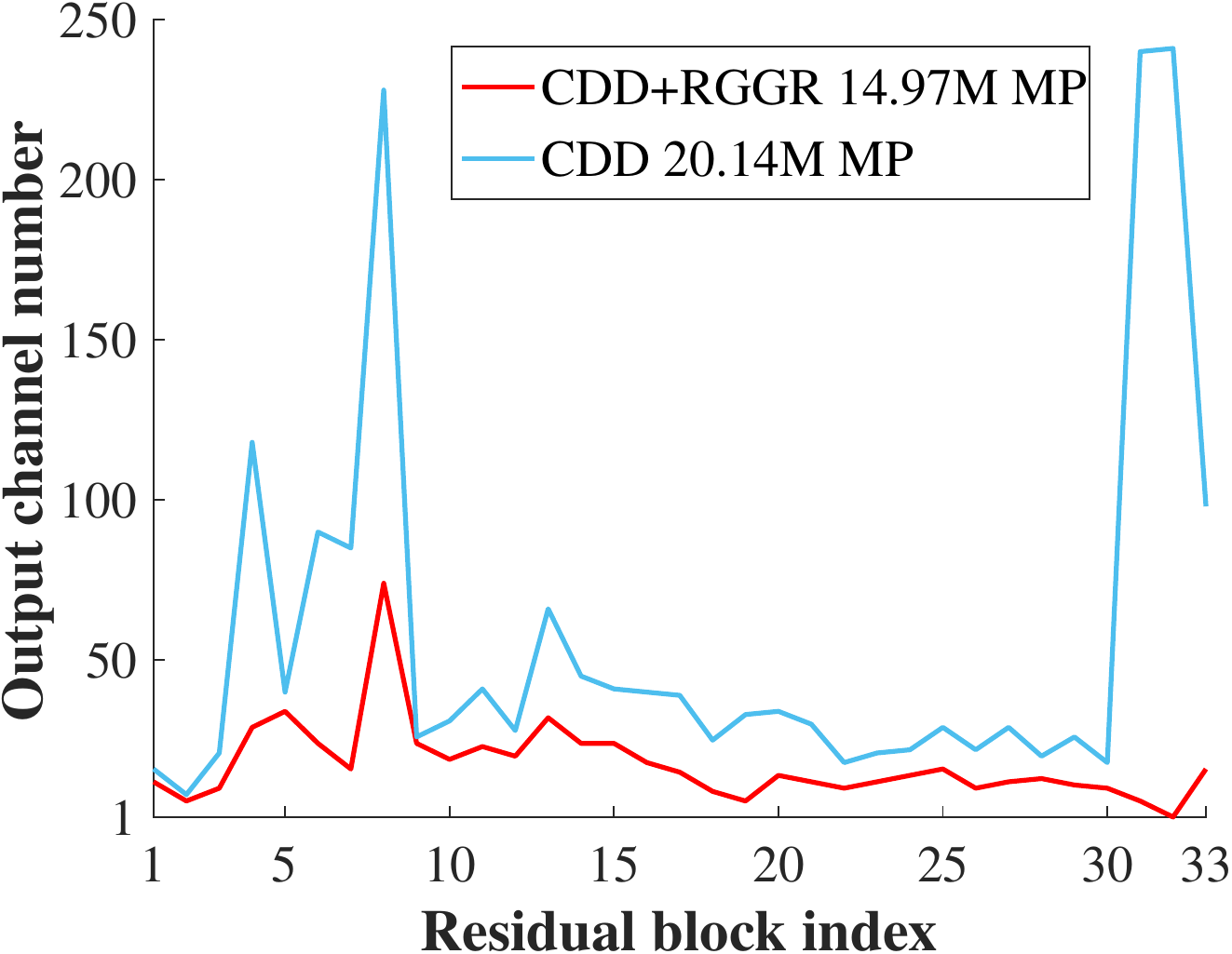}\vspace{-1mm}
		\caption{The output channel number of DGC in each residual block on In-shop \cite{inshop}. With RGGR, the output channel number of DGC is significantly reduced, while mAP only drops 0.1\%.}\label{fig:channel_number}
		\vspace{-2mm}
	\end{figure}

\tb{MSMT17 dataset.}
  From Table \ref{tab:sota_msmt17}, it is evident that the CDD+RGGR method using the ResNet101 model as the teacher network exhibits significant advantages in terms of accuracy performance metrics when compared to the non-compressed approaches, namely, IANet and BINet. Specifically, CDD+RGGR achieves the highest mAP (i.e., 60.98\%) and the highest R1 (i.e., 81.68\%) among all the methods. Additionally, it is worth noting that CDD+RGGR also outperforms the KD methods by 6.54M MP and 2.05G FLOPs in terms of inference performance, which further underscores its superiority as a compression method.

\begin{table}[tp]
	\caption{Evaluation of two stage cascaded designs on MSMT17.\vspace{-2mm}}\label{tab:two}
	\newcommand{\tabincell}[2]{\begin{tabular}{@{}#1@{}}#2\end{tabular}}
	\renewcommand\arraystretch{1}
	\centering
	\setlength{\tabcolsep}{3.5pt}
	%	\footnotesize
	\small
	\begin{threeparttable}
	\scalebox{1}{
		\setlength{\tabcolsep}{1pt}
		\begin{tabular}{c  cccc}
			\hlinewd{1.5pt}
			Methods &MP (M) & FLOPs (G)  &mAP (\%)     & R1 (\%)\\
			\hlinewd{0.8pt}
            ResRep \cite{resrep}  &14.85 &3.59  &51.17 &74.61 \\
            Self-KD + ResRep \cite{resrep} &14.82 &3.59 &56.17 &78.30\\
            % \hline
            % SSL \cite{lasso_prune} &15.11 &3.57  &46.23 &70.89\\
            % Self-KD + SSL \cite{lasso_prune} &14.84 &3.56 &51.02 &74.73\\
            % \hline
            CDD + RGGR  &15.00 &3.66 &60.98 &81.68  \\
			\hlinewd{1.5pt}
		\end{tabular}
	}
	
	\end{threeparttable}
\vspace{-.5cm}
\end{table}

\subsection{Analysis Experiments}
\subsubsection{The two stage method \textit{vs} CDD+RGGR} 
To evaluate the effectiveness of our integrated end-to-end KD+NP method, we conducted a comparative study by decomposing our approach into a cascaded setting. Specifically, we employ self-KD to distill a heavy student model, which is pruned to a light student by ResRep \cite{resrep}. Then, the light student is fine-tuned to restore some performance. Notably, in the two-stage cascaded method, our proposed RGGR retrieval-based metric is disabled, and we use the convolutional layer weight metric of ResRep \cite{resrep} instead for stable training. From Table \ref{tab:two}, CDD + RGGR significantly outperforms self-KD + ResRep by 4.18\% mAP and 3.38\% R1, which demonstrates that the integrated end-to-end KD+NP method is superior to the two-stage cascaded method.

\subsubsection{Hyper-parameter Analysis}
\noindent \textbf{The top-K retrieval result (i.e. $K$, in Eq. \eqref{eq:rk})} As $K$ value increases, RGGR has more reference information to zeroing the knowledge accumulation gradient of unimportant channels. From Fig. \ref{fig:K}, we can find that
the mAP performance of CDD+RGGR at $K > 1$ outperforms at $K=1$ on Veri776 \cite{veri}. Moreover, at $K > 1$, CDD+RGGR has good robustness on the mAP performance.  

% Besides, please refer to the supplementary material for the influence of $K$ on inference performance.

% The increase in K value in RGGR results in the availability of more reference information that helps zero the learning gradient of unimportant channels. As depicted in Figure 1, the impact of K on CDD+RGGR mAP is quite evident. It is observed that CDD+RGGR exhibits good robustness on mAP, and the mAP value at K > 1 surpasses that at K=1 on Veri776. Additionally, for further details on the impact of K on inference performance, please refer to the supplementary material.

\noindent \textbf{The channel selection mask ratio (i.e., $p (\%)$ in Eq. \eqref{eq:index})} Fig. \ref{fig:p} show that the impact of $p$ values on performance. Specifically, as the value of $p$ increases, the knowledge accumulation gradient of more channels becomes zero, leading to a slight decrease in mAP performance but an increase in inference performance for CDD+RGGR. For instance, increasing the $p$ value from $40\%$ to $70\%$ on In-shop \cite{inshop}, the mAP performance is reduced by 0.27\%. In contrast, the FLOPs performance improved from 4.9 G to 4.3 G.

\subsection{Qualitative Results}
Fig. \ref{fig:channel_number} shows the output channel number of slim DGC in each residual block on In-shop \cite{inshop}. We can find that RGGR can effectively sparse DGC and thus significantly reduce the output channel number. For example, in the $32$-th residual block, RGGR can reduce the output channel number of DGC from 241 to 1.

%Extensive experiments show that our method has superior inference speed and accuracy, e.g., on the VeRi-776 dataset, given the ResNet101 as a teacher, our method saves 67.13\% model parameters and 65.67\% FLOPs (around 24.13\% and 21.94\% higher than state-of-the-arts) without sacrificing accuracy(around 2.11\% mAP higher than state-of-the-arts).

\section{Conclusion}\label{sec:con}
This paper proposes a Capacity Dynamic Distillation framework (CDD) for efficient image retrieval. Specifically, CDD uses a heavy model as students to fully understand teachers' knowledge in early training.
Simultaneously, the student model is gradually compressed during the training by the DGC module. Furthermore, we propose the RGGR method to release the conflict between the learning gradient and the forgetting gradient. As a result, the heavy student model can be converted into a lightweight model without critical performance degeneration. Extensive experiments demonstrate that our method is superior to many state-of-the-art approaches.

\noindent\tb{Limitation.} Although our method achieves promising results on efficient CNNs, the performance on the transformer network is yet to be validated. In the future, we will extend our method to transformer-based knowledge distillation.

\noindent\tb{Broader Impact.} Our method demonstrates that the end-to-end aggregation of KD and NP helps construct large-capacity student models, which can inspire the community to continue to explore compression methods for large-capacity student models. In addition, our approach can be applied to learn the efficient model on other matching-dependent tasks (e.g., Object re-identification).

\noindent\tb{Acknowledgement.} The work is supported by Guangdong International Technology Coopertation Project (No.2022A0505050009); Key-Area Research and Development Program of Guangdong Province, China ( 2020B010165004,2020B010166003); National Natural Science Foundation of China (No. 61972162); Guangdong Natural Science Foundation (No. 2021A1515012625); Guangzhou Basic and Applied Research Project (No. 202102021074); and Guangdong Natural Science Funds for Distinguished Young Scholar (No. 2023B1515020097).
%%%%%%%%% REFERENCES
{\small
\bibliographystyle{ieee_fullname}
\bibliography{egbib}
}

\end{document}